%% file: 190_camera-ready.tex
\documentclass[runningheads,dvipsnames]{llncs}
\usepackage[T1]{fontenc}
\usepackage{graphicx}
\usepackage{float}

\usepackage[binary-units=true]{siunitx}
\usepackage{tikz}
\usetikzlibrary{trees, fit, matrix, positioning, patterns, shapes, shadows, shapes.arrows, arrows.meta, shapes.multipart, decorations.pathreplacing, calc, tikzmark, shapes.geometric}
\usetikzlibrary{arrows,intersections}
\usetikzlibrary{decorations.fractals}
\usetikzlibrary{shapes.geometric}
\usetikzlibrary{positioning}
\usetikzlibrary{calc}
\usepackage{pgfplotstable}
\usepackage{todonotes}
\usepackage{booktabs}

\usepackage{hyperref}
\usepackage{color}

\usepackage{amsmath}
\newcommand{\norm}[1]{\left \lVert #1 \right \rVert}

\begin{document}
\title{Less Memory Means smaller GPUs: Backpropagation with Compressed Activations}
\titlerunning{Backpropagation with Compressed Activations}
\author{Daniel Barley\inst{1}\orcidID{0009-0006-1607-1876} \and
Holger Fr\"oning\inst{1}\orcidID{0000-0001-9562-0680}}
\authorrunning{D. Barley \and H. Fr\"oning}
\institute{Computing Systems Group, ZITI, Heidelberg University, Heidelberg, Germany\\
\email{\{daniel.barley,holger.froening\}@ziti.uni-heidelberg.de}}
\maketitle              % typeset the header of the contribution
\begin{abstract}
% The abstract should briefly summarize the contents of the paper in

	The ever-growing scale of deep neural networks (DNNs) has lead to an equally
	rapid growth in computational resource requirements. Many recent
	architectures, most prominently Large Language Models, have to be trained
	using supercomputers with thousands of accelerators, such as GPUs or TPUs.
	Next to the vast number of floating point operations the memory footprint of
	DNNs is also exploding. In contrast, GPU architectures are notoriously short
	on memory. Even comparatively small architectures like some
	\emph{EfficientNet} variants cannot be trained on a single consumer-grade GPU
	at reasonable mini-batch sizes. During training, intermediate input
	activations have to be stored until backpropagation for gradient calculation.
	These make up the vast majority of the memory footprint. In this work we
	therefore consider compressing activation maps for the backward pass using
	pooling, which can reduce both the memory footprint and amount of data
	movement. The forward computation remains uncompressed. We empirically show
	convergence and study effects on feature detection at the example of the
	common vision architecture \emph{ResNet}. With this approach we are able to
	reduce the peak memory consumption by 29\% at the cost of a longer training
	schedule, while maintaining prediction accuracy compared to an uncompressed
	baseline.

\end{abstract}
\input{01-introduction}

\input{02-bgandrw}

\input{03-characterization}

\input{04-compression}

\input{05-experiments}

\input{06-discussion}

\begin{credits}
\subsubsection{\ackname}
This work is part of the "Model-Based AI" project,
which is funded by the Carl Zeiss Foundation.
\subsubsection{\discintname}
The authors have no competing interests to declare that are
relevant to the content of this article.
\end{credits}


\bibliographystyle{splncs04}
\bibliography{bibliography}
\end{document}

%% file: 01-introduction.tex
\section{Introduction}\label{sec:introduction}

The exponential growth of DNNs over the recent years in both parameter count
and computational complexity has sparked many advances in GPU technology and
vice versa. GPUs meanwhile feature special matrix multiply and most recently
even transformer accelerators\footnote{NVIDIA H100 transformer engine
\url{https://developer.nvidia.com/blog/nvidia-hopper-architecture-in-depth/}}.
These advancements of course allow for even larger DNNs, forming a feedback
loop. One of the major bottlenecks in this race is memory. GPU architectures
are notoriously short on memory, which is in direct conflict with the
ever-growing parameter counts. Different forms of pruning, quantization, and
compression of DNNs have therefore become active fields of research.

Pruning aims to lower the memory complexity by moving parts of a network toward
zero, which can then be encoded using a sparse format or dropped entirely. For
weights, a loss penalty term can be used to encourage more zero values for
example. The granularity of pruning can range from individual
weights~\cite{prechelt_connection_1997} to channels~\cite{he_channel_2017} to
whole layers of a network~\cite{schindler_parameterized_2020}. Pruning is of
course not limited to weights, but can also be applied to
gradients~\cite{tang_communicationefficient_2023} or
activations~\cite{barley_compressing_2023}.

Quantization on the other hand aims to reduce the memory footprint and
computational cost by reducing the bit width of parameters. This includes for
example half-precision floating point
representations~\cite{micikevicius_mixed_2018} and even integer
types~\cite{hengrui_efficient_2021,benoit_quantization_2018}. In the most
extreme cases networks are binarized~\cite{qin_binary_2020}, replacing the more
complex integer or floating point arithmetic with bitwise operations.

Pruning and quantization are not mutually exclusive of course. As both can have
serious implications on model accuracy and
performance~\cite{roth_resource_2024}, finding configurations that maintain
accuracy while improving performance is a non-trivial task. Automated
approaches can help to navigate the vast search
space~\cite{krieger_towards_2022}.

A commonality among most of these approaches is them focusing on model
inference. Because model inference is often performed on devices orders of
magnitude less powerful than the hardware used for training said model this is
a necessity of course. This even extends to architectural design. Architectures
like \emph{MobileNet}~\cite{howard_mobilenets_2017} and
\emph{EfficientNet}~\cite{tan_efficientnet_2019} make heavy use of weight
sharing and separable convolutions. While only a few hundred megabytes in size
during forward inference, training these models produces several gigabytes of
intermediate activations needed for backpropagation, more than is available on
most consumer-grade or embedded devices. In fact, as shown later in this work,
activations are responsible for the vast majority of the memory footprint
across architectures during training. The previous example might only be a
problem for hobbyists, but scaling to larger architectures, like language
models in particular, the memory demands affect even research facilities.
Additionally, lowering the memory requirements may enable online learning or
fine tuning on smaller devices like the NVIDIA Jetson platforms for
example\footnote{\url{https://developer.nvidia.com/embedded-computing}}.

To address this issue, we propose a method to compress stored activations
during the training process. We use pooling to reduce the dimensions of stored
activations and thereby also their memory footprint. In theory this can also
reduce the number of memory transactions during gradient computation in the
backward pass. To take full advantage, a custom operator is needed however,
which is still in development. The forward pass remains unchanged. We only
apply the pooling immediately before storing the activation for
backpropagation. This ensures an accurate loss computation. Also, unlike sparse
approaches, a pooled activation does not require any additional encoding that
causes overhead except for the original shape.

We test the effects of compressed activations on convergence and memory on
\emph{ResNet}~\cite{kaiming_deep_2016}. Additionally, we analyze the gradient
alignment between a regularly trained model and a compressed one to obtain
layer-wise sensitivity information.

In summary, this work makes the following contributions:
\begin{enumerate}
	\item We characterize various neural architectures regarding their memory
		footprint, demonstrating the amount of memory consumed by activations.
	\item We propose a method to reduce activation space by pooling blocks of
		activations based on averaging blocks of activations.
	\item We assess the implications of the proposed methods for different block
		sizes on ResNet. Experiments include observed loss curves for different
		compression configurations, reporting achieved memory size reduction, as
		well as trading among less memory size and an increased number of epochs to
		maintain test accuracy.
\end{enumerate}

%% file: 02-bgandrw.tex
\section{Related Work}

There exist similar approaches to employ compression or sparsity in DNNs,
optimizing for different metrics.

ReSprop~\cite{goli_resprop_2020} uses pruning, quantization, and compression of
activation maps during inference to accelerate the model and increase their
representational power. This is crucial in real-time tasks and mobile deployment.
However, this requires a pre-trained model of course.

Dithered Backprop~\cite{weidemann_dithered_2020} similarly uses a pruning and
compression scheme, but on preactivation gradients to induce sparsity and additionally
quantize non-zero values to low bit widths. Again, the goal is to achive a speedup
of the backward pass and to reduce computations, not memory. Additionally, the sparsity
induced by Dithered Backprop is unstructured, which is only compatible with GPUs
at extreme levels of sparsity.

Another very similar approach is presented in SWAT~\cite{raihan_swat_2020}.
SWAT prunes activations during training based on magnitude after a dense
forward computation with the goal of reducing FLOPs on sparse accelerators and
thereby accelerate the training. During forward propagation SWAT additionally
masks weights to that end. As with Dithered Backprop, the theoretical savings
do not translate well to execution on a GPU.

In a previous work~\cite{barley_compressing_2023}, partly based on SWAT, we
used structured pruning of activations with the explicit goal of reducing the
memory footprint during training. Although the encoding overhead shrinks with
increasing block size of the sparse structure, there is a non-negligible
overhead decoding the sparse structure in terms of computation and especially
memory movements. The regularity of the pooling operation presented here
eliminates the decoding overhead entirely.

%% file: 03-characterization.tex
\section{Model Characterization}

In this section we characterize the training memory footprint for various
vision architectures and analyze properties of the activation tensors saved. We
use a custom memory allocator to trace allocations in the
\texttt{torchvision}\footnote{\url{https://pytorch.org/vision/stable/index.html}}
implementations of said architectures for detailed analysis. For peak values it
is also possible to query PyTorch's default
allocator\footnote{\url{https://pytorch.org/docs/stable/generated/torch.cuda.max_memory_allocated.html}}.

\begin{figure}[t]
	\centering
	\includegraphics[width=\textwidth]{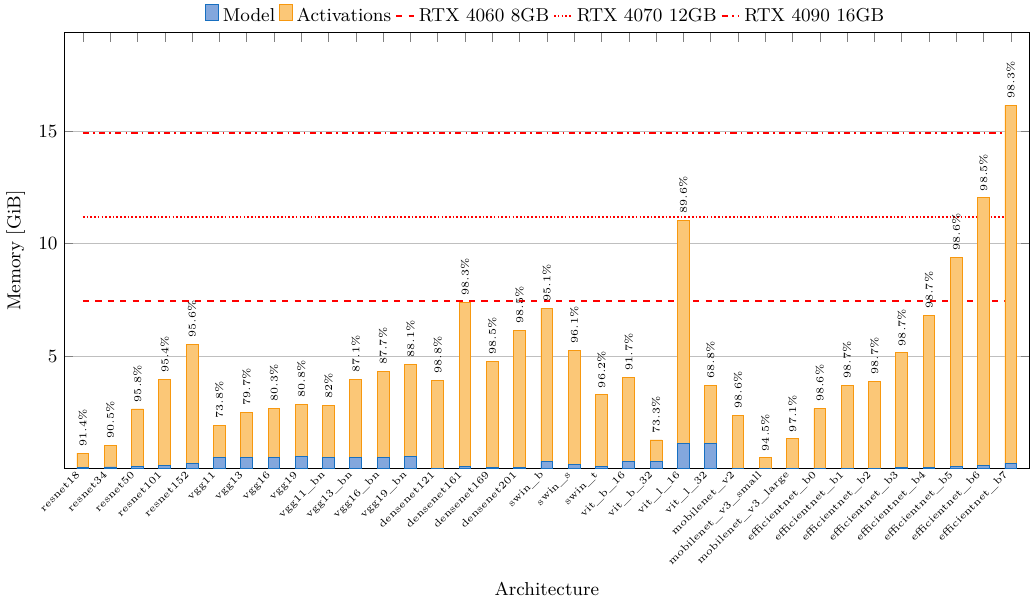}
	\caption{Memory footprint during training of common vision architectures
	split into model parameters and activations. The ratio is heavily skewed
	toward activations. The most balanced is \texttt{vit\_l\_32} at 68.8\%
	activations. On average, activations are responsible for 91.8\% of the memory
	consumed. \texttt{EfficientNet\_[B1-B4]} are the most extreme examples at
	98.7\% activations. The mini-batch size is 32 for all models. The red lines
	show the memory capacity of consumer-grade GPUs of the NVIDIA 4000 series for
	comparison. Note that the memory capacity is given in \si{\giga\byte}, while
	our measurements are in \si{\gibi\byte}. The printed percentages represent
	the proportion of activations in regard to the peak memory
	usage.}\label{fig:characterization:footprint}
\end{figure}

As mentioned in Section~\ref{sec:introduction}, stored activations are
responsible for most of the total memory consumed during training.
Figure~\ref{fig:characterization:footprint} compares the activation state to
the model parameter state for various prominent vision architectures.
Activations, on average, are responsible for 91.8\% of the total memory
consumed. The \emph{EfficientNet}\cite{tan_efficientnet_2019} architecture
shows the most extreme split. Due to weight sharing, convolutional layers
typically show a more skewed ratio than e.g. fully connected layers. On top of
that, EfficientNet makes heavy use of point-wise and depth-wise convolutions in
succession. This leads to 98.7\% of the memory being consumed by activations in
the case of \texttt{EfficientNet\_[B1-B4]}. In contrast, \emph{Vision
Transformers} (ViT)\cite{dosovitskiy_vit_2021} contain mainly fully-connected
components, and therefore show less skew. Although, at 68.8\% for
\texttt{vit\_l\_32} activations still make up more than two thirds of the
memory footprint. The activation footprint is also directly proportional to the
mini-batch size. 32 is a fairly low value for most of the architectures listed.
So the ratios shown here can be seen as best case values. Increasing mini-batch
size skews the ratio further toward activations.

\begin{figure}
	\centering
	\includegraphics[width=\textwidth]{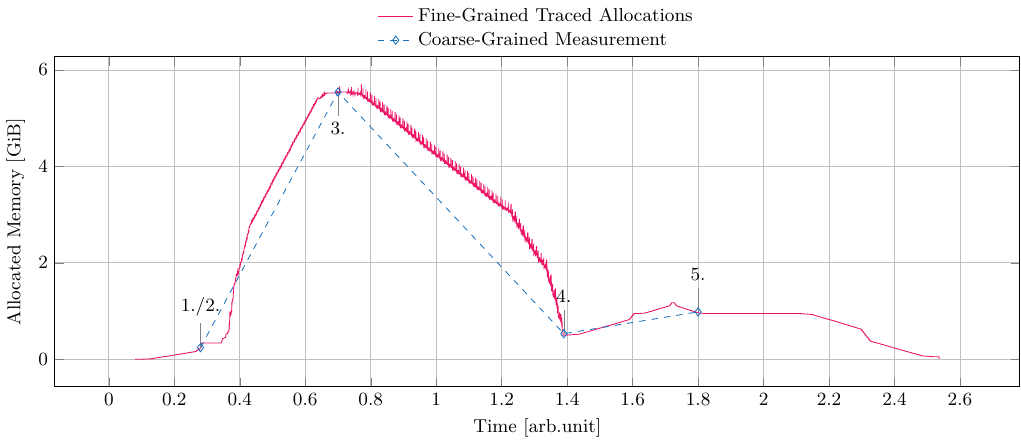}
	\caption{Fine-grained tracing of allocations during model initialization and
	processing of a single mini batch of data for the \texttt{ResNet152}
	architecture. The red line shows the step-wise allocation of forward
	activations and the reverse as the backward computation unfolds. After the
	backward computation is completed, a second peak can be observed during the
	optimization stage. We can simplify this representation by only measuring at
	points of interest: model/input (1./2.) initialization, forward peak (3.),
	after backward (4.), optimizer peak (5.) represented by the blue
	marks.}\label{fig:characterization:fine-grained}
\end{figure}

The previous analysis only shows the peak memory consumption after the forward
computation. However, allocation and deallocation of activations and other data
is dynamic. Additionally, during the following optimization stage there may be
additional memory allocations depending on the optimizer. A more detailed
analysis is therefore needed to assess the potential for reduction of the
memory footprint. Figure~\ref{fig:characterization:fine-grained} shows the
traced cumulated memory allocations at the example of the
\texttt{ResNet152} architecture at mini-batch size 32. Based on the
shape of the curve we can identify 5 points of interest that can be observed
instead of tracing all allocations.

\begin{enumerate}
	\item Model initialization: PyTorch allocates and initializes buffers for the
		model parameters (weights and biases).
	\item Input data initialization: Allocate a tensor for an input mini-batch.
	\item Forward pass peak: After the forward pass is completed memory usage is
		highest, as all activations have to be kept up to this point.
	\item Backward pass complete: During backpropagation saved activations are
		successively deallocated, leading to a minimum once all gradients are
		calculated.
	\item Optimizer peak: During the optimization pass additional memory is
		required for momenta for example. This highly depends on the optimizer
		used. In this example, \texttt{Adam}.
\end{enumerate}

%% file: 04-compression.tex
\section{Compressing Activation Maps}\label{sec:compression}

To address the problems related to encoding overhead for sparse data
structures~\cite{schindler_parameterized_2020}, we pursue a slightly different
approach. Instead of encoding non-zero values we use average pooling to reduce
the size of the activation maps. This has the substantial advantage that
besides the pooling kernel size no additional encoding is needed, allowing us
to drastically reduce the memory requirements.

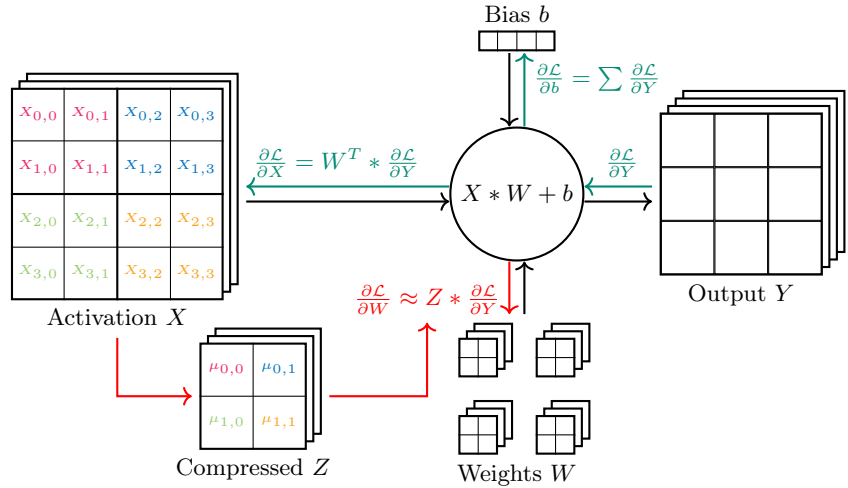
\begin{figure}[tb]
	\centering
	\input{mean}
	\caption{Example of activation map compression for a convolutional layer.
	During the forward computation the input activation $X$ is convolved with the
	weight tensor $W$ and a bias $b$ is added to produce output $Y$. Before
	saving the activation map for the backward pass, we compress it using
	pooling. Except for the original dimensions, this requires no additional
	encoding overhead. During backpropagation we inflate the compressed
	activation $Z$ back to the original dimensions and use it to obtain the weight
	gradient, as shown in red. The other gradients, shown in green, are not
	affected by this, as they do not depend on the activations. This ensures that
	the propagated activation gradient remains accurate throughout the network.
	Only the weight updates themselves are
	imprecise.}\label{fig:compression:mean}
\end{figure}

Figure~\ref{fig:compression:mean} shows the approach in detail.
Forward-propagated activations are not compressed. We only apply pooling to
input \(X\) after the output \(Y\) of the current layer is computed, yielding a
compressed activation \(Z\). By doing so, we always obtain an accurate loss
value. The pooled activation \(Z\) is stored until it is used to calculate the
weight gradient during backpropagation. With the kernel size of the pooling
kernel we can now inflate the activation back to its original dimensions and
calculate the weight gradient normally. We are currently working on a custom
implicit convolution kernel to elide the inflation and reduce memory
transactions in addition to memory capacity. 

In Figure~\ref{fig:compression:mean} one can also observe how the computation
of downstream gradients is changed based on the approximation of \(X\) by
\(Z\), e.g. by averaging. As denoted by the green color, both the downstream
gradients $\frac{\partial\mathcal{L}}{\partial X}$ and bias gradients
$\frac{\partial\mathcal{L}}{\partial b}$ are not affected by the compression.
Thus, the propagated downstream gradient remains accurate and therefore does
not accumulate errors through backpropagation. Only the weight updates
$\frac{\partial\mathcal{L}}{\partial W}$ become less accurate.

While in principle various forms of pooling are feasible, in this initial work
we start with average pooling. Similarly, for more diversity in weight updates,
one might change the pooling layout adaptively during training. Last, activation
pooling could also be selectively disabled on a per-layer basis.

%% file: mean.tex
\begin{tikzpicture}
	\matrix [
		draw,
		thick,
		matrix of math nodes, 
		double copy shadow={shadow xshift=1mm, shadow yshift=1mm},
		inner sep=0pt,
		fill=white,
		nodes={outer sep=0pt, minimum size=7mm, font=\tiny},
		label=below:{Activation $X$}
	] (activation)
	{
		|[WildStrawberry]| X_{0, 0} & |[WildStrawberry]| X_{0, 1} & |[RoyalBlue]| X_{0, 2} & |[RoyalBlue]| X_{0, 3} \\
		|[WildStrawberry]| X_{1, 0} & |[WildStrawberry]| X_{1, 1} & |[RoyalBlue]| X_{1, 2} & |[RoyalBlue]| X_{1, 3} \\
		|[YellowGreen]| X_{2, 0} & |[YellowGreen]| X_{2, 1} & |[YellowOrange]| X_{2, 2} & |[YellowOrange]| X_{2, 3} \\
		|[YellowGreen]| X_{3, 0} & |[YellowGreen]| X_{3, 1} & |[YellowOrange]| X_{3, 2} & |[YellowOrange]| X_{3, 3} \\
	};

	\node [draw, thick, circle, right=3cm of activation] (conv) {$X \ast W + b$};

	\matrix [
		matrix of nodes,
		thick,
		inner sep=0pt,
		nodes={fill=white, draw, minimum size=5mm, double copy shadow={shadow xshift=1mm, shadow yshift=1mm}},
		row sep = 5mm,
		column sep = 5mm,
		label=below:{Weights $W$},
		below=1cm of conv,
	] (weights)
	{
		{} & {} \\
		{} & {}\\
	};

	\matrix [
		draw,
		thick,
		matrix of math nodes, 
		double copy shadow={shadow xshift=1mm, shadow yshift=1mm},
		inner sep=0pt,
		fill=white,
		nodes={outer sep=0pt, minimum size=7mm, font=\tiny},
		label=below:{Compressed $Z$},
		left=2cm of weights,
	] (compressed)
	{
		|[WildStrawberry]| \mu_{0, 0} & |[RoyalBlue]| \mu_{0, 1} \\
		|[YellowGreen]| \mu_{1, 0} & |[YellowOrange]| \mu_{1, 1} \\
	};

	\node [draw, thick, above=1cm of conv, minimum height=2.5mm, minimum width=1cm, label=above:{Bias $b$}] (bias) {};

	\matrix [
		draw,
		thick,
		matrix of nodes,
		copy shadow={shadow xshift=3mm, shadow yshift=3mm},
		copy shadow={shadow xshift=2mm, shadow yshift=2mm},
		copy shadow={shadow xshift=1mm, shadow yshift=1mm},
		inner sep=0pt,
		fill=white,
		nodes={draw, thin, minimum size=7mm},
		label=below:{Output $Y$},
		right=1cm of conv,
	] (output)
	{
		{} & {} & {} \\
		{} & {} & {} \\
		{} & {} & {} \\
	};

	\draw [->, thick, shorten <= 3mm] ([xshift=1mm] weights.north) -- ([xshift=1mm] conv.south);
	\draw [<-, thick, shorten <= 3mm, red] ([xshift=-1mm] weights.north) -- ([xshift=-1mm] conv.south) node [midway, left] (gradw)
	{$\frac{\partial \mathcal{L}}{\partial W} \approx Z \ast \frac{\partial \mathcal{L}}{\partial Y}$};

	\draw [<-, thick, shorten <= 3mm, PineGreen] ([xshift=1mm] bias.north) -- ([xshift=1mm] conv.north) node [midway, right]
	{$\frac{\partial \mathcal{L}}{\partial b} = \sum \frac{\partial \mathcal{L}}{\partial Y}$};
	\draw [->, thick, shorten <= 3mm] ([xshift=-1mm] bias.north) -- ([xshift=-1mm] conv.north);

	\draw [<-, thick, shorten <= 3mm, PineGreen] ([yshift=1mm]activation.east) -- ([yshift=1mm]conv.west) node [midway, above]
	{$\frac{\partial \mathcal{L}}{\partial X} = W^T \ast \frac{\partial \mathcal{L}}{\partial Y}$};
	\draw [->, thick, shorten <= 3mm] ([yshift=-1mm]activation.east) -- ([yshift=-1mm]conv.west);

	\draw [<-, thick, shorten >= 1mm, PineGreen] ([yshift=1mm]conv.east) -- ([yshift=1mm]output.west) node [midway, above]
	{$\frac{\partial \mathcal{L}}{\partial Y}$};
	\draw [->, thick, shorten >= 1mm] ([yshift=-1mm]conv.east) -- ([yshift=-1mm]output.west);

	\draw (compressed.north) -- (compressed.south);
	\draw (compressed.west) -- (compressed.east);

	\draw [thick] (activation.north) -- (activation.south);
	\draw [thick] (activation.west) -- (activation.east);

	\draw ($(activation.north)!0.5!(activation.north east)$) -- ($(activation.south)!0.5!(activation.south east)$);
	\draw ($(activation.north)!0.5!(activation.north west)$) -- ($(activation.south)!0.5!(activation.south west)$);

	\draw ($(activation.north east)!0.5!(activation.east)$) -- ($(activation.north west)!0.5!(activation.west)$);
	\draw ($(activation.south east)!0.5!(activation.east)$) -- ($(activation.south west)!0.5!(activation.west)$);

	\draw ($(bias.north)!0.5!(bias.north east)$) -- ($(bias.south)!0.5!(bias.south east)$);
	\draw ($(bias.north)!0.5!(bias.north west)$) -- ($(bias.south)!0.5!(bias.south west)$);
	\draw (bias.north) -- (bias.south);

	\draw (weights-1-1.north) -- (weights-1-1.south);
	\draw (weights-1-1.west) -- (weights-1-1.east);

	\draw (weights-1-2.north) -- (weights-1-2.south);
	\draw (weights-1-2.west) -- (weights-1-2.east);

	\draw (weights-2-1.north) -- (weights-2-1.south);
	\draw (weights-2-1.west) -- (weights-2-1.east);

	\draw (weights-2-2.north) -- (weights-2-2.south);
	\draw (weights-2-2.west) -- (weights-2-2.east);

	\draw [red, thick, ->, shorten >= 1mm, shorten <= 5mm] (activation.south) |- (compressed.west);
	\draw [red, thick, ->, shorten <= 3mm] (compressed.east) -| (gradw.south);
\end{tikzpicture}

%% file: 05-experiments.tex
\section{Experiments}

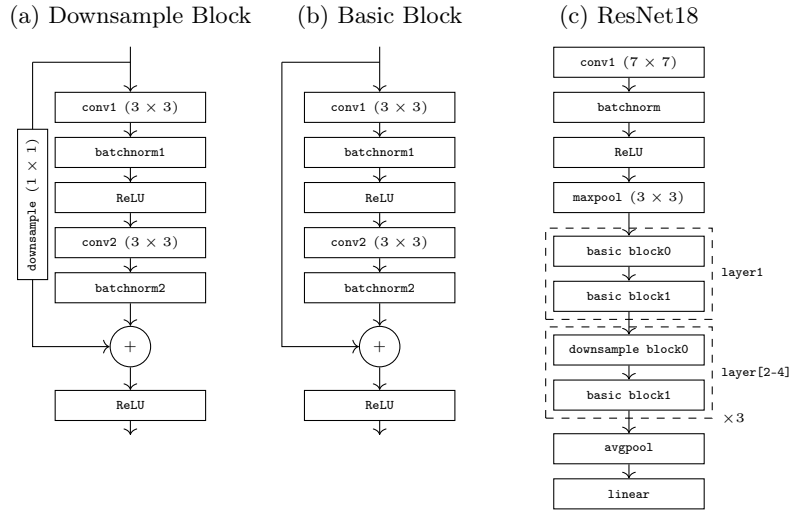
\begin{figure}
	\centering
	\input{resnet}
	\caption{Overview of the ResNet architecture (c) and its residual building
	blocks. The downsample block (a) features a $(1 \times 1)$ convolution in the
	residual path, whereas the basic block (b) adds the input
	unaltered.}\label{fig:experiments:resnet}
\end{figure}

We test the effects of our compression approach in terms of memory consumption
and accuracy on \texttt{ResNet18}. We use the operator described in
Section~\ref{sec:compression} to compress all activation maps throughout the
network evenly. The stem convolution \texttt{conv1} as well as linear
classifiers and batch normalization layers are kept unchanged.
Figure~\ref{fig:experiments:resnet} shows an overview of the architecture.
We train on the ImageNet dataset using SGD and a step learning rate scheduler.
The rest of the hyperparameters is kept close to the literature, except the
mini-batch size, which we set to 64.

\subsection{Training dynamics}

\begin{figure}[t]
	\centering
	\includegraphics[width=\textwidth]{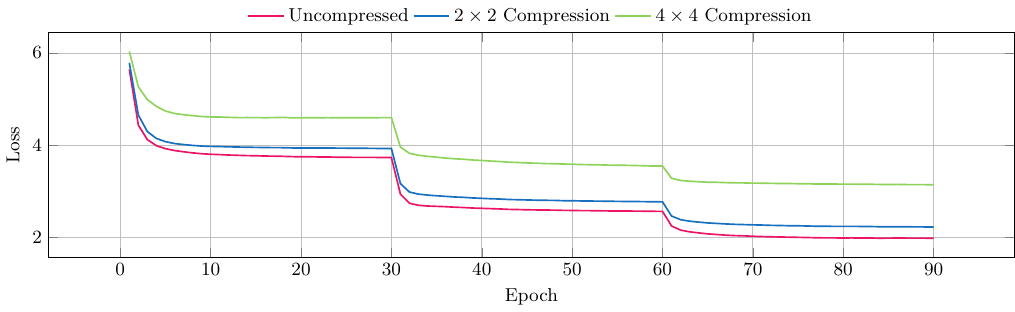}
	\caption{Training loss for \texttt{ResNet18} trained on ImageNet using SGD
	with momentum and step learning rate scheduler. Compression seems to
	introduce an offset to the loss curve.}\label{fig:experiments:loss}
\end{figure}

Figure~\ref{fig:experiments:loss} shows the effects of compression on the
training loss of ResNet. Compressing activation maps does not alter the trend
of the loss curve, but only introduces an offset. This leads us to believe a
longer training schedule might help to recover some of the accuracy lost by the
compression. After 90 Epochs accuracies of 67.7\%, 64\%, and 50.1\% are achieved
at the mentioned levels of compression shown.

\subsection{Layer sensitivity}\label{sec:experiments:layersensitivity}

To gain insight into the effects on an individual layer basis we compare the
cosine similarity of the weight gradients after one epoch to a model trained
without compression.
Cosine similarity is formally defined as follows:

\begin{equation}
	\cos (\textbf{a},\textbf{b}) = \frac{\textbf{a} \cdot \textbf{b}}{\norm{\textbf{a}} \norm{\textbf{b}} }
\end{equation}

A similarity of one means that gradients point in the same
direction exactly, while one of minus one means the opposite. 
Thus, here a higher value is better. 

\begin{figure}
	\centering
	\includegraphics[width=\textwidth]{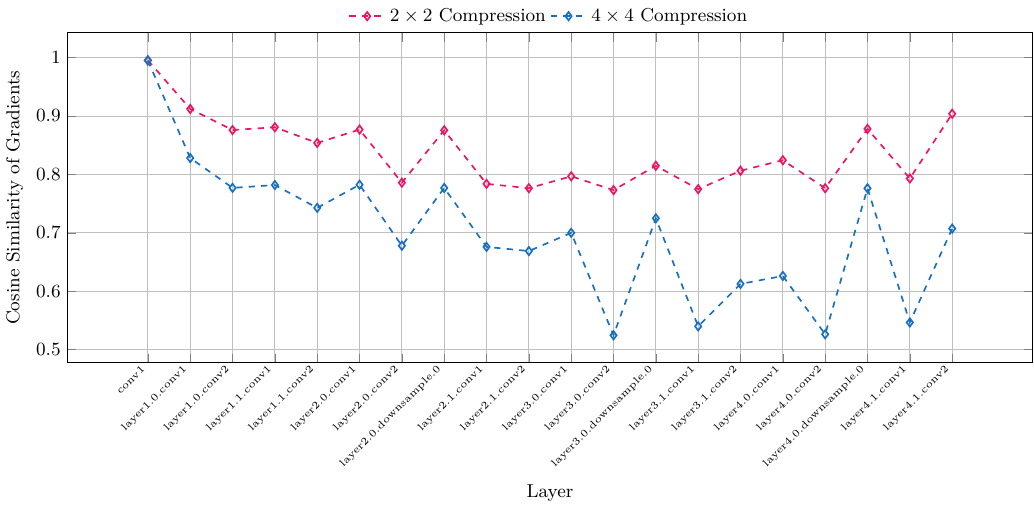}
	\caption{Capturing the cosine similarity of the weight gradients of a
	\texttt{ResNet18} after one epoch compared to an uncompressed model shows
	that not all layers are affected equally by compression. Downsample layers
	seem to be the least affected, while \texttt{conv2} seems to be especially
	sensitive in sublayer zero and \texttt{conv1} in sublayer
	one.}\label{fig:experiments:cosine}
\end{figure}

For this experiment we have fixed the data loader, so samples are processed by
all models in the same order. Due to the stochastic nature of the optimizer,
parameters diverge after the first epoch rendering a comparison meaningless.
Figure~\ref{fig:experiments:cosine} shows the similarities for
\texttt{ResNet18} for brevity. However, the other variants show very similar
trends. First, observing the uncompressed stem convolution \texttt{conv1} shows
that the gradient is indeed unaffected, as shown in
Figure~\ref{fig:compression:mean}. Next, we can see that the effects are not
uniform across layers. In the first sublayer \texttt{conv1} deviates further
from the baseline, while in the second sublayer it is \texttt{conv2}. This may
be a result of the gradient flow around the skip connections. Downsample layers
seem to be more resilient to compression. This is most likely due to the
reduced kernel size of $(1 \times 1)$.

\subsection{Memory footprint reduction}

Because compression does not introduce significant encoding overhead, we are
able to drastically lower the memory consumed by activations.
Table~\ref{tab:experiments:memory} shows the memory consumed by both weights
and activations for all convolutional layers in ResNet18. Additionally, the
table also shows the consumption for $(2 \times 2)$ and $(4 \times 4)$
compressed activations. As expected, pooling reduces the memory requirement by
$\frac{3}{4}$ and $\frac{15}{16}$ respectively, which is amounts to roughly
\SI{191}{\mebi\byte} and \SI{239}{\mebi\byte}.

With regard to the total memory footprint, the other layers have to be taken
into account as well. At around \SI{670}{\mebi\byte} total activation state,
$(2 \times 2)$ compression of the convolutional layer's activations results in
a 29\% reduction and $(4 \times 4)$ in a 36\% reduction of the overall memory
footprint.

\begin{table}
	\centering
	\caption{Memory consumption per convolutional layer with respect to
	parameters $W$, activations $X$, and compressed versions $Z$. To keep
	consistent with Figure~\ref{fig:characterization:footprint} a mini-batch size
	of 32 is assumed here.}\label{tab:experiments:memory}
	\begin{tabular}{lc|ccc}
		\toprule
		Layer & $W$ [\si{\mebi\byte}] & $X$ [\si{\mebi\byte}] & $Z$ $(2 \times 2)$ [\si{\mebi\byte}] & $Z$ $(4 \times 4)$ [\si{\mebi\byte}] \\
		\midrule
		\texttt{conv1} & 0.04 & 18.38 & 4.59 & 1.15 \\
		\texttt{layer1.0.conv1} & 0.14 & 24.50 & 6.12 & 1.53 \\
		\texttt{layer1.0.conv2} & 0.14 & 24.50 & 6.12 & 1.53 \\
		\texttt{layer1.1.conv1} & 0.14 & 24.50 & 6.12 & 1.53 \\
		\texttt{layer1.1.conv2} & 0.14 & 24.50 & 6.12 & 1.53 \\
		\texttt{layer2.0.conv1} & 0.28 & 24.50 & 6.12 & 1.53 \\
		\texttt{layer2.0.conv2} & 0.56 & 12.25 & 3.06 & 0.77 \\
		\texttt{layer2.0.downsample.0} & 0.56 & 24.50 & 6.12 & 1.53 \\
		\texttt{layer2.1.conv1} & 0.56 & 12.25 & 3.06 & 0.77 \\
		\texttt{layer2.1.conv2} & 1.12 & 12.25 & 3.06 & 0.77 \\
		\texttt{layer3.0.conv1} & 2.25 & 12.25 & 3.06 & 0.77 \\
		\texttt{layer3.0.conv2} & 2.25 & 6.12 & 1.53 & 0.28 \\
		\texttt{layer3.0.downsample.0} & 2.25 & 12.25 & 3.06 & 0.77 \\
		\texttt{layer3.1.conv1} & 4.50 & 6.12 & 1.53 & 0.28 \\
		\texttt{layer3.1.conv2} & 9.00 & 6.12 & 1.53 & 0.28 \\
		\texttt{layer4.0.conv1} & 9.00 & 6.12 & 1.53 & 0.28 \\
		\texttt{layer4.0.conv2} & 9.00 & 3.06 & 0.56 & 0.06 \\
		\midrule
		Sum & 41.94 & 254.19 & 63.34 (-75\%) & 15.35 (-94\%) \\
		\bottomrule
	\end{tabular}
\end{table}

\subsection{Maintaining accuracy}

To restore some of the lost accuracy due to compression, experiments with more
training epochs were conducted. The goal is to find a viable trade-off between
the increased end-to-end training time and the reduced memory consumption.
Extending the training schedule from 90 to 120 epochs indeed shows improvements,
especially in the case of $(2 \times 2)$ compression. We are able to come within
1.3\% of the baseline accuracy, which we view as acceptable, considering the
fluctuations based on weight initialization being only slightly smaller.
$(4 \times 4)$ compression results are still unacceptable.

%% file: resnet.tex
\begin{tikzpicture}[
	node distance=6mm,
	transform shape,
	every node/.style={
		font=\tiny\ttfamily,
		text=black,
		minimum width=2cm,
		minimum height=4mm,
	}
]
\begin{scope}
	\node [draw](conv1) {conv1 $(7 \times 7)$};
	\node [draw, below of = conv1] (bn1) {batchnorm};
	\node [draw, below of = bn1] (relu) {ReLU};
	\node [draw, below of = relu] (maxpool) {maxpool $(3 \times 3)$};

	\node [draw, color=black, text=black, below = 3mm of maxpool] (block0) {basic block0};
	\node [draw, color=black, text=black, below of = block0] (block1) {basic block1};
	\node [draw, fit=(block0)(block1), dashed, inner sep=1mm] (layer0) {};
	\node [right, minimum width=0pt, minimum height=0pt] at (layer0.east) {layer1};

	\node [draw, color=black, text=black, below = 3mm of block1] (block2) {downsample block0};
	\node [draw, color=black, text=black, below of = block2] (block3) {basic block1};
	\node [draw, fit=(block2)(block3), dashed, inner sep=1mm] (layer1) {};
	\node [right, minimum width=0pt, minimum height=0pt] at (layer1.south east) {$\times 3$};
	\node [right, minimum width=0pt, minimum height=0pt] at (layer1.east) {layer[2-4]};

	\node [draw, below = 3mm of block3] (avgpool) {avgpool};
	\node [draw, below of = avgpool] (fc) {linear};

	\draw [->] (conv1) -- (bn1);
	\draw [->] (bn1) -- (relu);
	\draw [->] (relu) -- (maxpool);
	\draw [->] (maxpool) -- (block0);
	\draw [->] (block0) -- (block1);
	\draw [->] (block1) -- (block2);
	\draw [->] (block2) -- (block3);
	\draw [->] (block3) -- (avgpool);
	\draw [->] (avgpool) -- (fc);
\end{scope}

	\begin{scope}
	\node [left = 1.3cm of conv1] (basic_input) {};
	\node [draw, below of = basic_input] (basic_conv1) {conv1 $(3 \times 3)$};
	\node [draw, below of = basic_conv1] (basic_bn1) {batchnorm1};
	\node [draw, below of = basic_bn1] (basic_relu) {ReLU};
	\node [draw, below of = basic_relu] (basic_conv2) {conv2 $(3 \times 3)$};
	\node [draw, below of = basic_conv2] (basic_bn2) {batchnorm2};
	\node [draw, circle, below = 3mm of basic_bn2, minimum width=0pt, minimum height=0pt] (basic_res) {$+$};
	\node [draw, below = 3mm of basic_res] (basic_relu_out) {ReLU};
	\node [below of = basic_relu_out] (basic_output) {};

	\draw [->] (basic_input.north) -- (basic_conv1);
	\draw [->] (basic_conv1) -- (basic_bn1);
	\draw [->] (basic_bn1) -- (basic_relu);
	\draw [->] (basic_relu) -- (basic_conv2);
	\draw [->] (basic_conv2) -- (basic_bn2);
	\draw [->] (basic_bn2) -- (basic_res);
	\draw  (basic_input.center)+(-1.3,0) -- (basic_input.center);
	\draw [->] (basic_input.center)+(-1.3,0) |- (basic_res.west);
	\draw [->] (basic_res) -- (basic_relu_out);
	\draw [->] (basic_relu_out) -- (basic_output);
\end{scope}

\begin{scope}
	\node [left = 1.3cm of basic_input] (downsample_input) {};
	\node [draw, below of = downsample_input] (downsample_conv1) {conv1 $(3 \times 3)$};
	\node [draw, below of = downsample_conv1] (downsample_bn1) {batchnorm1};
	\node [draw, below of = downsample_bn1] (downsample_relu) {ReLU};
	\node [draw, below of = downsample_relu] (downsample_conv2) {conv2 $(3 \times 3)$};
	\node [draw, below of = downsample_conv2] (downsample_bn2) {batchnorm2};
	\node [draw, circle, below = 3mm of downsample_bn2, minimum width=0pt, minimum height=0pt] (downsample_res) {$+$};
	\node [draw, below = 3mm of downsample_res] (downsample_relu_out) {ReLU};
	\node [below of = downsample_relu_out] (downsample_output) {};

	\draw [->] (downsample_input.north) -- (downsample_conv1);
	\draw [->] (downsample_conv1) -- (downsample_bn1);
	\draw [->] (downsample_bn1) -- (downsample_relu);
	\draw [->] (downsample_relu) -- (downsample_conv2);
	\draw [->] (downsample_conv2) -- (downsample_bn2);
	\draw [->] (downsample_bn2) -- (downsample_res);
	\draw  (downsample_input.center)+(-1.3,0) -- (downsample_input.center);
	\draw [->] (downsample_input.center)+(-1.3,0) |- node [near start, draw, rotate=90, fill=white] {downsample $(1 \times 1)$} (downsample_res.west);
	\draw [->] (downsample_res) -- (downsample_relu_out);
	\draw [->] (downsample_relu_out) -- (downsample_output);
\end{scope}

\node [above of = downsample_input, font=\small\rmfamily] {(a) Downsample Block};
\node [above of = basic_input, font=\small\rmfamily] {(b) Basic Block};
\node [above of = conv1, font=\small\rmfamily] {(c) ResNet18};

\end{tikzpicture}

%% file: 06-discussion.tex
\section{Discussion and Future Work}\label{sec:discussion}

This work is still in progress, but the preliminary results seem promising and
we intend to expand the concept further. Next steps include more architectures,
especially EfficientNet on the CNN side and Transformers like ViT and swin.
For EfficientNet in particular we expect good results, based on the observations
made on the $(1 \times 1)$ downsample convolutions. Additionally, we plan on
expanding compression to batch normalization layers as well, as they produce
activations with the same dimensions of the respective convolutional layer.
Non-uniform pooling kernel shapes could also be considered.

The results from Figure~\ref{fig:experiments:loss} show that a uniform $(4
\times 4)$ compression is not viable. However, with $(2 \times 2)$ compression
alone, we are able to reduce the peak memory usage by 29\%. Considering
Figure~\ref{fig:characterization:footprint}, a reduction by 29\% would, in many
cases, allow for the use of a "smaller" GPU, going from a 4090 to a 4080 for
example in the case of \texttt{EfficientNet\_B7}. For tasks like fine-tuning
and retraining even embedded devices like the Jetson Nano series become viable.
With the custom compressed backward convolution in place, we plan on running
experiments on a Jetson Orin Nano device including tasks like retraining and
online learning in the future.

As yet, the process is also fairly invasive. Custom implementations of the
network architectures are required. We therefore plan to integrate the
compression operators in to our structured sparse operator library
\texttt{deep-sparse-nine} for a more streamlined workflow. This would also
allow to easily mix compression and sparsity.

Regarding the methodology, cosine similarity data could be used to
automatically find a viable per-layer configuration, perhaps even
adaptively during training via control epochs. At the cost of additional
training time, one could initialize an uncompressed model to repeat the cosine
similarity analysis shown in subsection~\ref{sec:experiments:layersensitivity}
mid-training. Based on the data, individual layers could be updated to a new
kernel size or trained uncompressed for a period of time.

Our Code is available on GitHub: \url{https://github.com/danielbarley/cnn_mean_compression}.